%% file: paper.tex
\documentclass[10pt,twocolumn,letterpaper]{article}

\usepackage{iccv}
\usepackage{times}
\usepackage{epsfig}
\usepackage{graphicx}
\usepackage{amsmath}
\usepackage{amssymb}
\usepackage{float}

\usepackage[pagebackref=true,breaklinks=true,letterpaper=true,colorlinks,bookmarks=false]{hyperref}

\iccvfinalcopy 

\def\iccvPaperID{1097} 

\makeatletter
\def\Demo{%
  \def\Ginclude@graphics##1{%
      \rule{\@ifundefined{Gin@@ewidth}{150pt}{\Gin@@ewidth}}%
      {\@ifundefined{Gin@@eheight}{100pt}{\Gin@@eheight}}}}
\makeatother

\newcommand\blfootnote[1]{%
  \begingroup
  \renewcommand\thefootnote{}\footnote{#1}%
  \addtocounter{footnote}{-1}%
  \endgroup
}

\ificcvfinal\fi
\begin{document}

\title{A deep architecture for unified aesthetic prediction}

\author{Naila Murray and Albert Gordo\\
Naver Labs Europe\\
{\tt\small naila.murray@naverlabs.com}
}

\maketitle

\begin{abstract}
Image aesthetics has become an important criterion for visual content curation on social media sites and media content repositories.
Previous work on aesthetic prediction models in the computer vision community has focused on aesthetic score prediction or binary image labeling.
However, raw aesthetic annotations are in the form of score histograms and provide richer and more precise information than binary labels or mean scores.
Consequently, in this work we focus on the rarely-studied problem of predicting aesthetic score distributions and propose a novel architecture and training procedure for our model.
Our model achieves state-of-the-art results on the standard AVA large-scale benchmark dataset for three tasks: (i) aesthetic quality classification; (ii) aesthetic score regression; and (iii) aesthetic score distribution prediction, all while using one model trained only for the distribution prediction task. We also introduce a method to modify an image such that its predicted aesthetics changes, and use this modification to gain insight into our model.\blfootnote{Work done while at XRCE (now Naver Labs Europe). AG now at Facebook.}
\end{abstract}

\input{intro}

\input{related}
\input{method}

\input{exps}

\input{conclusion}

\clearpage

{\small
\bibliographystyle{ieee}
\bibliography{aesthetics}
}

\end{document}

%% file: intro.tex
\section{Introduction}

As the quantity of images in both online and offline repositories continues to rapidly grow, properties beyond the semantic content of the images are increasingly used to both cull content and retrieve relevant items from databases.
The aesthetic properties of images, in particular, have received increased attention in the computer vision community in recent years \cite{DJLW06,Ke2006,Luo2008,obrador2010role,dhar2011,Luo2011,joshi2011aesthetics,SanPedro2012,obrador2012towards}.

Most early works proposed designing hand-crafted features that correlate to known aesthetic principles such as the rule-of-thirds, or using complementary colors \cite{DJLW06,Luo2008,li2010towards,simond2015image}.
More recent works have used either generic unsupervised visual representations \cite{Marches2011,marchesotti2015discovering,simond2015image} or used supervised deep architectures to generate appropriate representations \cite{lu2014rapid,lu2015deep,jin2016image,kong2016photo,mai2016composition,zhou2016joint}.
These representations have typically been used to train models for predicting either binary labels corresponding to high or low aesthetic quality \cite{DJLW06,Luo2008,Marches2011,mai2016composition,zhou2016joint} or for predicting an aesthetic score on a scale e.g. from 1 to 10 \cite{DattaW10,SanPedro2012,kong2016photo,jin2016image}.

\begin{figure}[!t]
 \centering
 \centering
 \includegraphics[width=\linewidth]{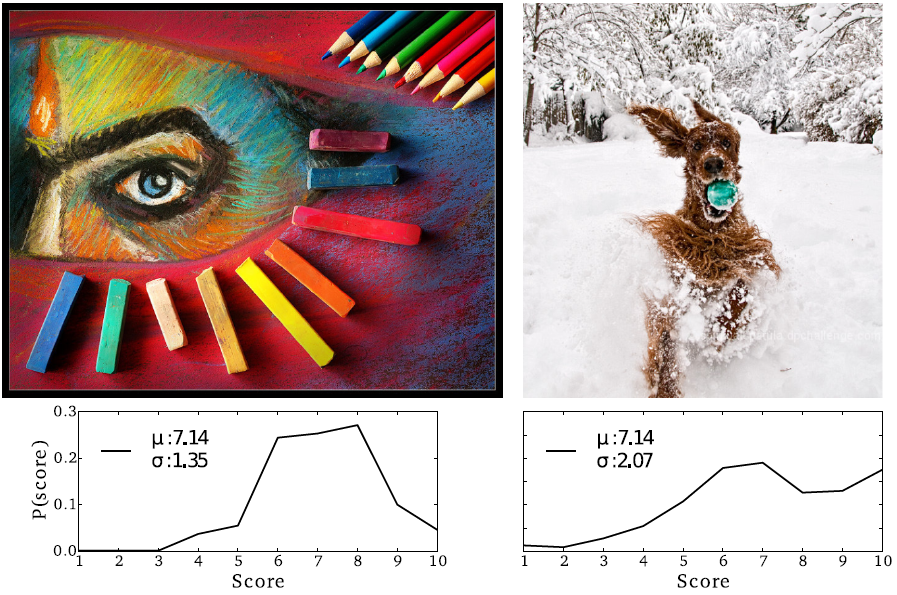}%
 \caption{Two test images from the AVA dataset \cite{murray2012ava}. Both images have the same mean score, but very different score distributions: the left image has a unimodal distribution while the right image has a bimodal one, indicating a divergence of opinion.}
 \label{fig:eg_dists}  
 \end{figure}

However, because aesthetic quality is a subjective property, aesthetics annotations rarely come in the form of a binary tag or a single score, but usually form a distribution of scores (e.g., from 1-5 stars in Flickr, from 1-10 in \texttt{www.dpchallenge.com}, or from 1-7 in \texttt{www.Photo.net}).
Binary labels are typically derived from such distributions by computing the mean of the distribution of scores given to an image and then thresholding the mean score.
However, a binary label or mean score removes information about the distribution of opinions about the image, which can indicate, for example, how much consensus of opinion exists, or whether there are widely divergent points of view (see Figure~\ref{fig:eg_dists}).
The few works which have aimed to predict aesthetic score distributions have either focused on mitigating unreliable scores from annotators \cite{Wu2011}, or on mitigating the dataset bias \cite{jin2016image}.

 In this work we focus on the design and training of a convolutional neural network (CNN) architecture which predicts aesthetic score distributions. Most deep aesthetic models fine-tune a network that was initially trained to classify heavily downsampled and/or cropped images from the ImageNet database. However, as noted by Mai \etal \cite{mai2016composition}, maintaining high resolution and the original aspect ratio is critical for preserving information about an image's aesthetics. Therefore, we propose a model which accepts high-resolution images of an arbitrary aspect ratio, and which normalizes effectively the features which are pooled from different-sized image regions.

In our method, we also use pre-trained convolutional layers that were initially trained with fixed-size input images from ImageNet. We propose to retrain the convolutional layers of this model to adapt the layers to the statistics of the arbitrarily-sized input images, while maintaining its semantic discriminative power. To achieve this, we learn to predict soft object classification targets provided by a ``teacher network''.
In this we take inspiration from model distillation \cite{hinton2015distilling} and ``learning without forgetting'' \cite{li2016learning}. We show that these re-trained layers provide a better starting point from which to fine-tune the network for aesthetic prediction.

To gain an understanding of the image characteristics that our aesthetic model deems important, we also propose an approach inspired by adversarial example learning that modifies an image to either increase or decrease its predicted scores given a fixed model. By visualizing the image regions with the largest pixel value changes we gain insights into what parts of the image receive the attention of the model.

Our main contributions are the following:
\begin{enumerate}
\item We introduce a CNN architecture that goes beyond predicting only the mean score of the images and instead is trained to regress to the distribution of scores.
    The model employs the Huber loss, which is more robust to outliers;
\item We propose a fine-tuning strategy which better leverages privileged information about the semantic content of images;
\item We introduce a technique to modify an image such that its predicted aesthetic score distribution changes, and use this modification to provide insight into the type of visual content that tends to highly influence the aesthetics of an image.
\end{enumerate}

Our aesthetic prediction model achieves state-of-the-art results, by a large margin, for three tasks: distribution prediction, mean score prediction, and binary label prediction, by using one model. In particular, it shows a 2.1\% absolute improvement on classification accuracy, and a 27\% relative improvement for mean score prediction.

The remainder of the article is organized as follows: in section~\ref{sec:related}, we discuss related work. In section~\ref{sec:method}, we describe the architecture of, and training procedure for, our aesthetic prediction model. We provide an extensive experimental validation using the AVA dataset in section~\ref{sec:exps} and we conclude in section~\ref{sec:conclusions}.

%% file: related.tex
\section{Related work}\label{sec:related}
We discuss work related to visual representations for aesthetic prediction and training aesthetic prediction models.

\textbf{Visual representations}:
Since the publication of Datta \etal 's seminal work on the topic \cite{DJLW06}, a plethora of hand-crafted visual features for aesthetics prediction have been proposed \cite{DJLW06,dhar2011,Ke2006,Luo2011,Luo2008}. These features are explicitly designed to capture rules and techniques used by skilled photographers. Datta \etal proposed 56 visual features which could be extracted from an image. In particular, colorfulness features were extracted by comparing the distribution of an image's colors to a reference distribution, and average pixel intensity was used to represent light exposure. Several other features related to familiarity, texture, size, aspect ratio, region composition, low depth-of-field, and shape convexity were also designed by \cite{DJLW06}. Later works proposed similar representations. In particular, those proposed by Ke \etal \cite{Ke2006} were designed to describe the spatial distribution of edges, color distribution, simplicity, blur, contrast and brightness. Luo \& Tang \cite{Luo2008} first segmented the subject region of an image before extracting from it high-level semantic features related to composition, lighting, focus controlling and color. Dhar \etal \cite{dhar2011} aimed to describe image aesthetics using compositional, content-related and illumination-related attributes nameable by humans.\\[2mm]
Photographic techniques are often high-level and difficult to model, leading to a gap between the representational power of hand-crafted aesthetic representations and the aesthetic quality of an image. In \cite{Marches2011}, it was shown that generic image representations, \ie descriptors which were not designed for aesthetic image analysis, could significantly out-perform aesthetics-specific features. Specifically, the Bag-of-Visual-words descriptor~\cite{SZ03} and the Fisher vector~\cite{sanchez2013image} were used for aesthetic quality prediction. These generic descriptors are able to implicitly encode the aesthetic characteristics of an image by describing the distribution of intensity and color gradients in local image patches.\\[2mm]
Recently, deep methods have been proposed for learning representations specific to the aesthetic prediction task \cite{lu2014rapid,lu2015deep,jin2016image,kong2016photo,mai2016composition,zhou2016joint}
. These methods largely re-use pre-trained models that were developed for image classification, such as AlexNet \cite{KrSuHi12} and VGGNet \cite{Simonyan14c}, and adapt them to the task of aesthetic quality classification \cite{lu2014rapid,lu2015deep,mai2016composition,zhou2016joint}, or regression \cite{jin2016image,kong2016photo}.

\textbf{Training aesthetic prediction models}:
Most works on aesthetic quality prediction aim to minimize classification error, by training an SVM or a logistic regressor \cite{lu2014rapid,lu2015deep,mai2016composition,zhou2016joint}.
In \cite{jin2016image,kong2016photo}, a Euclidean loss is used to minimize regression error, while in \cite{kong2016photo,murray2012learning}, ranking losses were used.
To our knowledge there are only two previous works which attempt to predict aesthetic score distributions \cite{Wu2011,jin2016image}. In \cite{Wu2011}, loss functions were proposed which aimed to adapt the model updates to the reliability of the distribution.
The mean scores of images in aesthetics datasets in general, and the AVA dataset in particular, follow a normal distribution. As a consequence, there are few examples of images with very high or very low scores. As a result, aesthetic prediction models tend to perform poorly for such images. In \cite{jin2016image}, the authors addressed this bias by re-weighting training samples in inverse proportion to the atypicality of their mean score. While this approach has the effect of better predicting the aesthetic quality of low- and high-quality images, it degraded the performance of the model for mediocre images. In the same work, a model was trained to predict aesthetic score distributions. However, the model was evaluated only on the task of predicting the mean and variance of the resultant distributions, rather than predicting the distributions themselves.
Several works have shown that incorporating semantic supervision helps when training aesthetic quality prediction models. For example, in \cite{mai2016composition}, including an auxiliary task of scene categorization boosted model performance on aesthetic quality classification. In \cite{kong2016photo}, in addition to aesthetic score regression, additional tasks of aesthetic ranking, aesthetic attribute prediction and semantic category prediction were shown to boost categorization and regression performance.
Our proposed model, described next, incorporates image semantics using a teacher network, and uses robust regression to train a network which predicts a score distribution for an image of arbitrary size and aspect ratio.



%% file: method.tex
\section{Our aesthetic prediction model}\label{sec:method}
Our proposed method aims to directly predict score distributions using a CNN trained end-to-end.
In what follows, we first describe the optimization objective of our model (Section \ref{sec:dist_pred}.
We then describe the main model architecture (Section \ref{sec:arch}). Finally, we discuss how to leverage privileged information to improve the accuracy of the model (Section \ref{sec:lsi}) and implementation details (Section \ref{sec:imp}).

\subsection{Score distribution prediction}\label{sec:dist_pred}
For aesthetic datasets, raw annotations are provided in the form of score histograms, where the $i$-th bin contains the number of voters that gave a discrete score of $i$ to the image. These histograms are $\ell_1$-normalized as a preprocessing step to obtain the empirical distributions.
These are then used to derive annotations for prediction tasks such as classification and mean score prediction. We denote the $d$-dimensional ground truth distributions with $\boldsymbol{g}$, with $d=10$ (scores from 1 to 10, both included) in all our experiments.

Our network is comprised of several convolutional layers followed by a series of fully-connected layers, of which the last produces the predicted distribution $\boldsymbol{p}$.
We cast our learning problem as a structured regression task where the regression layer predicts each discrete probability independently.
Aesthetic models which perform regression typically use the Euclidean loss as a loss function \cite{jin2016image,kong2016photo}.
However, the Euclidean loss is not robust to outliers, which poses a problem for aesthetics prediction, given that aesthetics is a subjective property and outlier opinions are likely.
Instead, we aim to minimize the Huber loss, which is known to be less sensitive to outliers \cite{huber1964robust}:
\begin{equation}
L_{\delta}(p_i, g_i)=
\begin{cases}
    \frac{1}{2}(p_i-g_i)^2,& \text{for $|p_j - g_j|\leq \delta$} \\
\delta(|p_i - g_i|-\frac{1}{2}\delta) & \text{otherwise},
\end{cases}
\end{equation}

\noindent where $p_i$ is the predicted value for score $i$, $g_i$ is the ground truth score, and $\delta=\frac{1}{\sigma^2}$ controls the degree of influence given to larger prediction errors.
In our experiments we use $\sigma=3$, chosen after cross-validation on a validation set. The loss between a predicted distribution and the ground truth is the sum of the losses of the individual predictions, \ie, $L_{\delta}(\boldsymbol{p}, \boldsymbol{g}) = \sum_{i=1}^d L_{\delta}(p_i, g_i)$.

\begin{figure*}[!ht]
\centering
\includegraphics[width=1\linewidth]{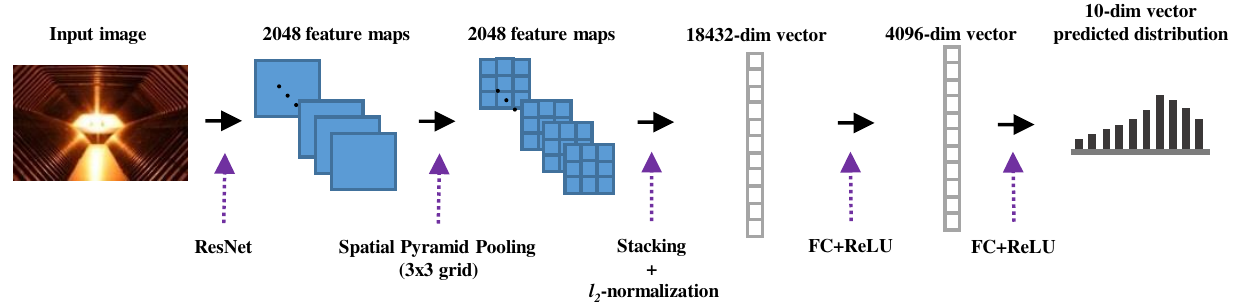}
\caption{Score distribution prediction pipeline: purple elements refer to operations. The convolutional layers of the ResNet-101 model are applied to the input image. The resultant response maps are adaptively pooled, and feature vectors from each cell and channel are stacked, resulting in a fixed-length representation which is $\ell_2$-normalized. Two fully-connected layers are then applied to this fixed-length vector, outputting the final predicted score distribution.}
\label{fig:pipeline}
\end{figure*}

\subsection{Network architecture}\label{sec:arch}
Our architecture is based on ResNet-101 \cite{He2015}. This network has been shown to outperform other popular deep architectures such as the Inception network \cite{szegedy2015going} and VGGNet \cite{Simonyan14c} for tasks such as image categorization \cite{szegedy2015going} and semantic segmentation \cite{dai2015instance}. However, we also report results using VGGNet as a base network to compare our approach with other deep aesthetic models, which typically use VGGNet.

Standard deep architectures for computer vision, including ResNet and VGGNet, assume a fixed-dimension input image.
As a result, images typically are either cropped or resized (or both) to conform to the expected dimensions.
However, as discussed in \cite{mai2016composition}, image composition, aspect ratio and resolution are critical aesthetic characteristics of an image, and are severely distorted by cropping or resizing.
To eliminate the need for resizing, one can trivially transform the fully-connected layers into convolutional ones, slide the network in a convolutional manner through the input image, and aggregate the responses through max-pooling or average-pooling to obtain a single output of the same dimensionality independently of the size of the image. However, such approaches are simply averaging / max-pooling the final predictions over the different regions of the image, and do not preserve the image composition.

Instead, we follow the approach introduced in \cite{he2014spatial} and used in \cite{mai2016composition}, and add an adaptive spatial pyramid pooling (SPP) layer after the last convolutional layer of the network.
The SPP layer is adaptive in the sense that the grid of the spatial pyramid is identical for each image, preserving the image composition, but the extent of each cell in the grid is resized on a per-image basis.
In our network we used a $n\times n$ grid, with $n=3$.
After pooling, the $C\times n^2$ feature vectors are stacked into a single vector of $Cn^2$ dimensions, where $C$ denotes the number of channels or response maps.
Importantly, this single vector is of fixed length for all training and testing images.
We also $\ell_2$-normalize the stacked vector to normalize response magnitudes across images.
This approach was also taken in \cite{gordo2016deep} in the context of image retrieval, and in \cite{bell2016inside} for object detection.
Finally, after the spatial pyramid pooling, we add two additional fully-connected (FC) layers followed by ReLU. The final layer outputs the 10-dimensional vector $\boldsymbol{p}$ which is the predicted distribution.
The architecture is illustrated in Figure~\ref{fig:pipeline}. Note that this setup allows one not only to test the model using high-resolution, non-cropped images, but also, and more importantly, to train with them.

\subsection{Leveraging semantic information}\label{sec:lsi}
Incorporating semantic information by learning an auxiliary task improves aesthetic prediction \cite{kong2016photo,mai2016composition}.
However, there are no large-scale image aesthetics datasets which also have large-scale semantic annotations.
While images from AVA do have tags (only some of which are semantic in nature), they are non-exhaustive (on the website, images are allowed at most two tags) and therefore the images are only partially annotated.
Kong \etal \cite{kong2016photo} addressed this issue by clustering the training images and using the cluster assignments as categorization labels for training a category predictor which shares low-level layers with the aesthetic branch of the network.
However, this makes the strong assumption that all of the clusters are semantically consistent.
Mai \etal \cite{mai2016composition}  add an additional deep network which was trained on a different image dataset which has no aesthetic labels but does have scene category labels.
The outputs of this model were aggregated with those of the aesthetic model to compute a final prediction.
However, this auxiliary network could not be fine-tuned for the aesthetic prediction task.

Here, we propose to leverage the semantic information that the original ResNet-101 \cite{He2015} model contains by virtue of having been trained on the ImageNet classification task using 1000 semantic categories: We aim at preserving this semantic information through the changes in domain, both in terms of the dataset and task domain shift and in terms of the architecture changes (higher resolution images and spatial pooling with SPP).
To this end we draw inspiration from model distillation \cite{hinton2015distilling} and ``learning without forgetting'' \cite{li2016learning} techniques.
We first use the original ResNet-101 classification model to obtain 1000-dimensional vectors from the AVA dataset images by taking the softmax output $\boldsymbol{s}$.
These vectors encode the predicted distribution of objects in the images, and provide relatively strong supervision on the semantic content of the AVA training images. 

We then use $\boldsymbol{s}$ as targets for training a modified version of the proposed architecture described in section~\ref{sec:arch}.
This architecture has one additional FC layer and softmax normalization after the output of the SPP that directly predicts the original distribution $\boldsymbol{s}$. 
The model is trained using the cross-entropy loss. By learning to predict $\boldsymbol{s}$, we are able to overcome the shift that happened when we changed the resolution of the input images as well as the changes in the architecture due to the SPP.
Ideally, this additional semantic loss should be learned simultaneously with the aesthetic prediction loss in a multi-task setting.
However, this approach did not bring noticeable improvements due to the difficulty of calibrating the two losses simultaneously.
Instead, we first perform the fine-tuning to predict ${s}$, \ie, to recover the semantic-prediction abilities of the model, and only then start predicting the aesthetic distribution. Empirically, this approach brought larger improvements.
Including this intermediate fine-tuning procedure adapts the convolutional layers of the model to changes in the statistics of the old and new representation caused by the new input image layer and SPP layer.

\subsection{Implementation details}\label{sec:imp}

We implemented our model using the Caffe library \cite{jia2014caffe} and optimized it using SGD with momentum. Training on the AVA dataset's approximately 250k training images took 2 weeks on a single Nvidia M40 GPU.
Although our network can train and evaluate with images of arbitrary dimensions, very large images drastically decrease training and evaluation speed and pose memory issues due to GPU memory constraints. Therefore, in practice we resize each image such that the smaller image dimension equaled 500, while maintaining the original aspect ratio. This resulted in significant loss of resolution in some cases, but is a significantly higher resolution than is typically used for convolutional networks.
We used a batch size of 128, a learning rate of $10^{-3}$, momentum of $0.9$ and weight decay of $5\cdot 10^{-4}$. We reduced the learning rate after every 20k iterations.
The convolutional layers were pre-trained on ImageNet \cite{imagenet}.

We initially included an $\ell_1$-normalization layer after the regression layer to normalize $\boldsymbol{p}$ to a valid distribution, but this strong constraint led to slow convergence during training and no accuracy gains. We noted a similar effect using a softmax normalization layer.
Our final model does not include a normalization step in the optimization, and we $\ell_1$-normalize $\boldsymbol{p}$ as a post-processing step during evaluation.
Note that due to the final ReLU layer, the predictions are always non-negative.

%% file: exps.tex

\section{Experimental results}\label{sec:exps}
We evaluate on the AVA dataset \cite{murray2012ava,marchesotti2015discovering}. This dataset contains more than 250000 images from \texttt{www.dpchallenge.com}, a website for photography enthusiasts, in which users rate and critique each other's photographs. Each image also has a histogram of scores from 1 - 10, given by DPChallenge members. Each image is rated by 210 members on average, resulting in a robust score histogram. We $\ell_1$-normalize this histogram to generate our target training and testing distributions.
We used the train/test split defined by the authors of \cite{murray2012ava} and held out 2000 images from the train set for validating hyper-parameters. The test set contains 20000 image IDs from \texttt{www.dpchallenge.com}, of which 19930 images were available at the time of download.

We evaluate our method with respect to three aesthetic quality prediction tasks: (i) aesthetic quality classification, (ii) mean aesthetic score prediction, and (iii) aesthetic score distribution prediction.
To construct the ground-truth for the mean aesthetic score prediction task, we compute the mean score from the score histogram of each test image.
For the classification task we threshold this mean score using a threshold $t$ to label the image as having low or high aesthetic quality.
At testing time, we derive our predicted mean $\mu$ based on the predicted, non-normalized distribution $\boldsymbol{p}$ as
\begin{equation}
    \mu=\sum^{10}_{i=1}i \cdot \hat{p}_i,
\label{eq:mup}
\end{equation}
where $\hat{\boldsymbol{p}}$ is the $\ell_1$-normalized $\boldsymbol{p}$.
We then threshold $\mu$ to derive the predicted binary labels.

\paragraph{Metrics:}
We report several metrics related to the three prediction tasks that we address:
\begin{itemize}
\item Image aesthetic quality classification: We report accuracy, \ie the percentage of correct labels. We binarize the ground-truth and predicted mean scores using a threshold of 5, as is standard practice for this dataset.
\item Image aesthetic mean score regression: We report Spearman's correlation coefficient $\rho$, and the mean squared error (MSE), both of which have been reported by different prior works.
\item Image aesthetic score distribution: We report the cumulative distribution loss (CDLoss), and the KL-divergence. The CDLoss was proposed by Wu \etal \cite{Wu2011} and is computed as:
\begin{equation}
CDLoss(\boldsymbol{p},\boldsymbol{g})=\sum_{i}(p^c_i - g^c_i)^2,
\end{equation}
where $\boldsymbol{p^c}$ and $\boldsymbol{g^c}$ are the cumulative distributions of the predicted ($\boldsymbol{p}$) and ground-truth ($\boldsymbol{g}$) distributions.
\end{itemize}

\subsection{Analysis of model components}\label{sec:models}

We compare 5 versions of our model:
\begin{itemize}
\item \texttt{VGG-ImN-mean}: The architecture of this model differs from the one shown in Figure~\ref{fig:pipeline} in two ways: (i) the convolutional layers of ResNet were replaced with those of VGG; and (ii) rather than predict ten values corresponding to a score distribution, the model predicts one value corresponding to the mean score. This model was trained with the Huber loss function, using mean scores calculated from the ground-truth distributions of the AVA images. The convolutional layers of the VGG model were pretrained on ImageNet.
\item \texttt{VGG-ImN-dist}: The architecture of this model is identical to that of VGG-mean, except for the final layer which outputs 10 predicted values corresponding to a score distribution.
\item \texttt{ResNet-ImN-dist}: The architecture of this model is illustrated in Figure~\ref{fig:pipeline}. The convolutional layers of the ResNet model were pretrained on ImageNet.
\item \texttt{ResNet-AVA-dist}: The architecture of this model is also that of Figure~\ref{fig:pipeline}. However, the convolutional layers were first fine-tuned on AVA to preserve the semantic prediction capabilities as described in section~\ref{sec:lsi}. We used an initial learning rate of $10^{-3}$ that was reduced by a factor of 10 every 20k iterations until convergence.
\item \texttt{ResNet-AVA-dist-m}: This is the same as \texttt{ResNet-AVA-dist}, but the model contains an additional branch after the last convolutional layer that performs global max-pooling, leading to a more global representation that complements the SPP. The global max-pooling is followed by $\ell_2$ normalization and by two fully-connected layers, of which the last one predicts the distribution.
    As the output vector of the global pooling has a relatively small dimensionality compared to the output of the spatial-pyramid pooling (2048 vs 18432), the model size does not increase significantly with the addition of this new branch.
    This is in contrast to methods such as that of Mai \etal \cite{mai2016composition}, which used different spatial pyramids, but trained separate networks for each of them, drastically increasing the overall size of their model.
    We initialized the model weights with the trained \texttt{ResNet-AVA-dist} model, and trained the two new FC layers with a learning rate of $10^{-3}$, while all other layers were trained with a reduced learning rate of $10^{-4}$.
At testing time, the predicted distributions of the two branches are combined with late fusion as a weighted averaged of the predictions, where the optimal weights are learned on the validation set.
\end{itemize}

 \begin{figure*}[!t]
 \centering
 \includegraphics[width=\textwidth]{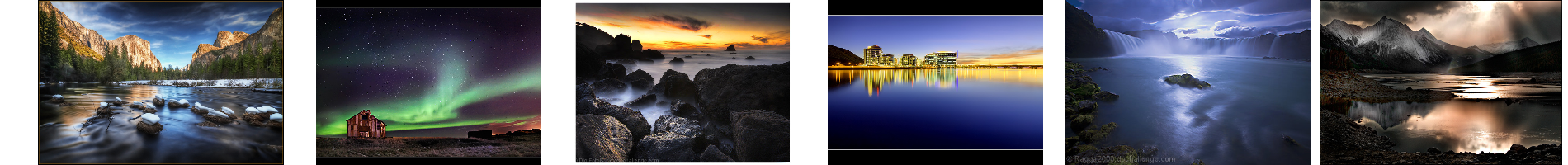}%

 \includegraphics[width=\textwidth]{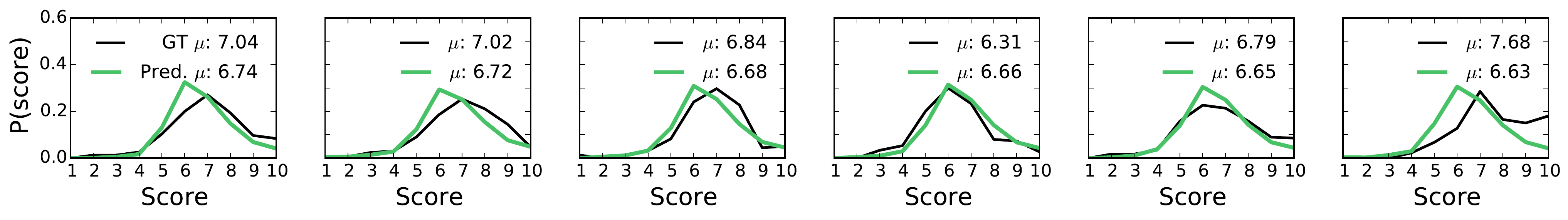}%
 
 \includegraphics[width=\textwidth]{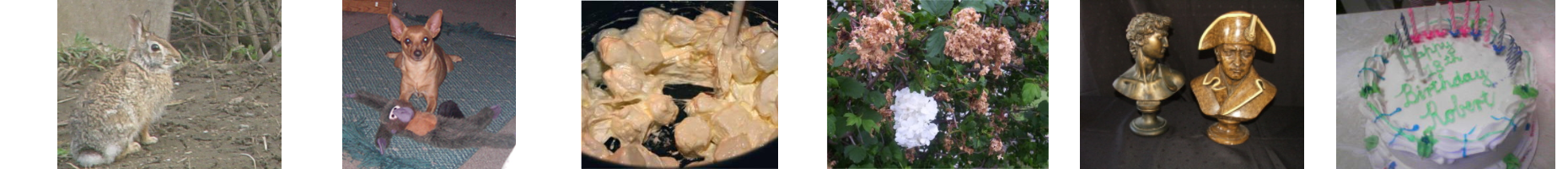}%

 \includegraphics[width=\textwidth]{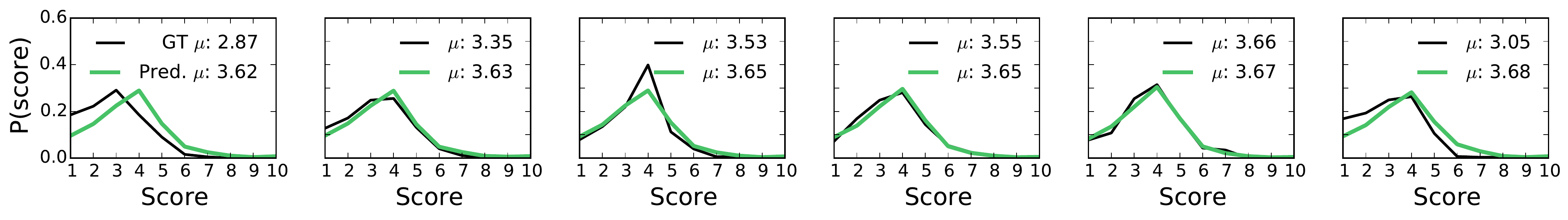}%

 \caption{Top 2 rows: The 6 most high-scoring images, as predicted by our APM model, coupled with plots of their ground-truth and predicted score distributions. Bottom 2 rows: The 6 most low-scoring images, as predicted by our APM model, coupled with plots of their ground-truth and predicted score distributions.}
 \label{fig:best_worst}  
 \end{figure*}

To compare the effectiveness of the Huber loss with that of the Euclidean loss, we also report results on a version of $\texttt{VGG-ImN-dist}$ trained with the Euclidean loss, termed $\texttt{VGG-ImN-dist-Eucl}$.
The quantitative results of these different versions are shown in Table~\ref{tab:apm_comp}.
We first note that \texttt{VGG-ImN-dist} performs on par with \texttt{VGG-ImN-mean} in terms of MSE, and shows a moderate improvement in $\rho$, even though it was not trained explicitly for the task of score prediction.
One reason for this may be that, as the last fully-connected layer independently predicts each $p_i$ in equation~\ref{eq:mup}, the model acts as an ensemble of predictors. Ensemble methods, including CNN ensembles, have been shown to reduce error due to having (at least partially) uncorrelated errors \cite{dietterich2000ensemble,KrSuHi12}.
Next, comparing the results for \texttt{VGG-ImN-dist-Eucl} and \texttt{VGG-ImN-dist} clearly shows the benefit of using a loss that is robust to outliers, and therefore more appropriate for the subjective nature of aesthetic appreciation.

We also observe a large improvement across all metrics when using pre-trained convolutional layers from ResNet-101 instead of from VGG (\texttt{VGG-ImN-dist} vs. \texttt{ResNet-ImN-dist}). This is unsurprising given ResNet's better classification, detection, and segmentation performances compared to VGGNet \cite{He2015}.
We therefore used ResNet convolutional layers for the remaining experiments.
Fine-tuning for classification on AVA, as described in section~\ref{sec:lsi}, resulted in moderate improvements in MSE and $\rho$, showing the benefit of adapting the convolutional layers in order to better represent the semantic content in an aesthetic prediction context.
The final modification of an additional prediction branch using globally-pooled images gives small improvements in general, but moderate improvement for image classification. 
We refer to our final aesthetic prediction model, \texttt{ResNet-AVA-dist-m}, as APM in the sequel.



\begin{table}
\setlength{\tabcolsep}{2pt}
\centering
{\footnotesize
\begin{tabular}{|l|c|c|c|c|c|}
\hline
Method & KLDIV $\downarrow$ & CDLoss $\downarrow$ & MSE $\downarrow$ & $\rho$ $\uparrow$ & Acc.(\%) $\uparrow$ \\
\hline
\hline
\texttt{VGG-ImN-mean} & - & - & 0.314 & 0.659 & 78.4 \\
\hline
\texttt{VGG-ImN-dist-Eucl.} & 0.134 & 0.0821 & 0.383 & 0.560 & 76.3 \\
\texttt{VGG-ImN-dist} & 0.128 & 0.0673 & 0.311 & 0.665 & 78.7 \\
\texttt{ResNet-ImN-dist} & 0.105 & 0.0618 & 0.285 & 0.702 & 79.7 \\
\texttt{ResNet-AVA-dist} & 0.104 & 0.0609 & 0.280 & 0.707 & 79.8 \\
\texttt{ResNet-AVA-dist-m} & \textbf{0.103} & \textbf{0.0607} & \textbf{0.279} & \textbf{0.709} & \textbf{80.3} \\
\hline
\end{tabular}
}
\caption{\label{tab:apm_comp} Results on the AVA testset for the proposed model, comparing different architecture choices and training procedures. Please refer to section~\ref{sec:models} for details.}
\end{table}

\subsection{Comparison to state-of-the-art methods}\label{sec:soa_comp}
Table~\ref{tab:soa_comp} compares several state-of-the-art aesthetic prediction models and our proposed APM model.
Jin \etal \cite{jin2016image} used a non-standard split of the AVA dataset, choosing to randomly select 5000 images from AVA for evaluation, and using the remaining 249530 images for training their models. As the official test set of AVA contains approximately 20000 randomly chosen images, the results are likely to be comparable. We report their results on a model trained and tested for score prediction (denoted by Jin \etal \cite{jin2016image}-Reg) and on one which was trained for score distribution prediction, but was evaluated on the task of predicting the mean and variance of the distribution (Jin \etal \cite{jin2016image}-HP).

Wu \etal \cite{Wu2011} evaluated their method on a dataset based on that of \cite{DJLW08}, and randomly split 9000 images collected from \texttt{www.dpchallenge.com} into training and testing sets. They repeated the random split 5 times and reported the average results. We were unable to reproduce these training splits\footnote{We contacted the authors of \cite{Wu2011} but they were unable to provide their lists of train/test splits}. However, given that the AVA test set also consists of randomly-chosen images from \texttt{www.dpchallenge.com} for testing, the results are also likely to be comparable. This is the only previous work, to our knowledge, that attempts to predict probability distributions.

Our simplest proposed model, \texttt{VGG-ImN-mean}, already outperforms the best reported MSE and $\rho$ for aesthetic score prediction by wide margins. Both prior works used models that incorporated pre-trained VGG layers, and are therefore comparable to our model in that regard.
The accuracy of \texttt{VGG-ImN-mean} for aesthetic quality classification is also marginally better than the best reported result for this task (Zhou \etal \cite{zhou2016joint}, which also incorporated pre-trained VGG layers), even though our model was trained for regression and not classification.
This is likely due to the fact that the aesthetic classification ground truth is inherently noisy, as it is based on a hard and  arbitrary thresholding of ground truth scores.
Training instead with the mean score provides a more robust supervision.

Our final APM model outperforms the state of the art on all tasks by wide margins.
In particular, it shows a 2.1\% absolute improvement on classification accuracy, and a 27\% relative improvement for mean score prediction (from 0.558 to 0.709).

\begin{table}
\centering
{\footnotesize
\begin{tabular}{|l|c|c|c|c|}
\hline
Method & CDLoss $\downarrow$ & MSE $\downarrow$ & $\rho$ $\uparrow$ & Acc.(\%) $\uparrow$ \\
\hline
\hline
Wu \etal \cite{Wu2011}$\dagger$ & 0.1061 & - & - & - \\
Jin \etal \cite{jin2016image}-HP $\dagger$ & - & 0.636 & - & -\\
Jin \etal \cite{jin2016image}-Reg $\dagger$ & - & 0.337 & - & -\\
Mai \etal \cite{mai2016composition} & - & - & - & 77.4 \\
Kong \etal \cite{kong2016photo} & - & - & 0.558 & 77.8 \\
Zhou \etal \cite{zhou2016joint} & - & - & - & 78.2 \\
\hline
\texttt{VGG-ImN-dist} & 0.0673 & 0.311 & 0.665 & 78.7 \\
\textbf{APM} & \textbf{0.0607} & \textbf{0.279} & \textbf{0.709} & \textbf{80.3} \\
\hline
\end{tabular}
}
\vspace{2pt}
\caption{ \label{tab:soa_comp} Comparison to state-of-the-art aesthetic prediction methods. $\dagger$ uses non-standard test set: please refer to section~\ref{sec:soa_comp} for details.}
\end{table}
 
Figure~\ref{fig:best_worst} shows the top 6 and bottom 6 images in the test set when ranked by predicted mean score (derived from the predicted distribution). Plots of the ground-truth and predicted distributions are also shown. The model is able to predict both the distribution itself, and the mean of the distribution to a high degree of accuracy, with almost perfect reconstruction in some cases. While it may appear that the model is biased towards images of landscapes, the database itself has a strong bias towards such photographs, as reported by Marchesotti \etal \cite{marchesotti2015discovering}.

\begin{figure}
 \centering
 \includegraphics[width=\linewidth]{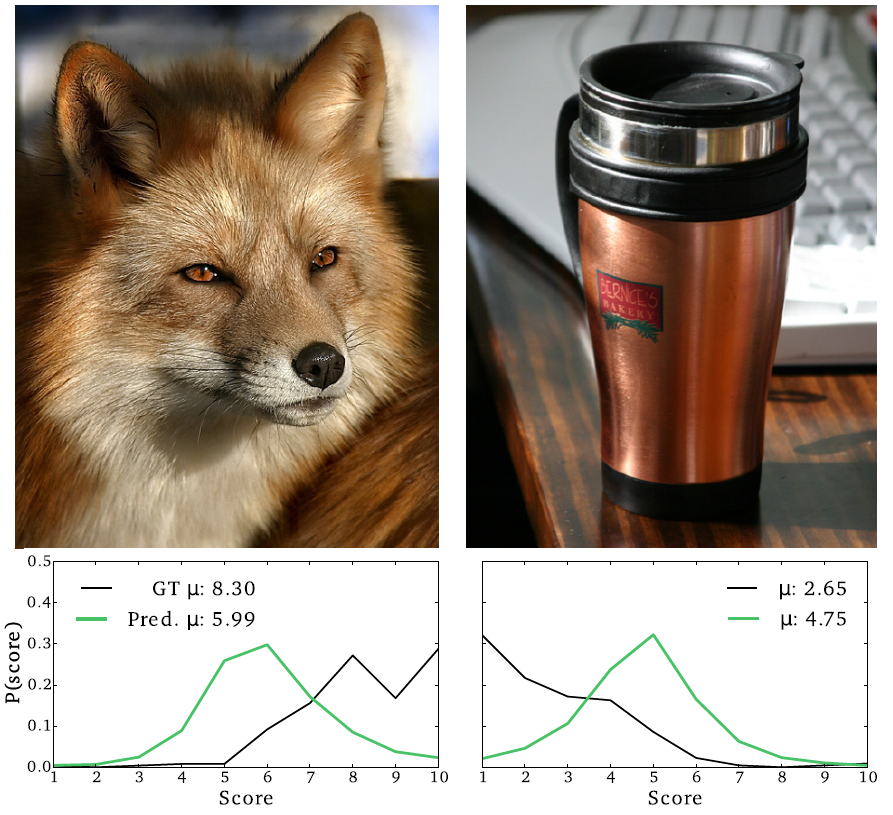}%
 \caption{Failure cases: APM performs poorly on very skewed distributions, which correspond to images that are either very high-rated (\eg image on the left) or very low-rated (\eg image on the right.}
 \label{fig:failure}  
 \end{figure}

In Figure~\ref{fig:failure}, we show two illustrative examples of the main failure modes of our model. As mentioned previously, aesthetics datasets in general, and AVA in particular, have relatively few very high- or very low-rated images. As a result, our trained APM model performs poorly on such images, which have very non-Gaussian distributions.

%
%
%
%

\subsection{Interpreting the model}
While deep models for aesthetic prediction show excellent quantitative performance, particularly when compared with shallow or unsupervised features, they lack interpretability.
Works such as \cite{kao2017deep,kao2016hierarchical} analysed visualizations of the learned  weights of the first convolutional layer of their proposed models. However, these weights reflect very localized and low-level information and thus are not very informative about their models' decisions. Some works have proposed methods to visualize and understand deep model decisions and representations in the context of classification (\cite{mahendran2015understanding}, \cite{zeiler2014visualizing}). However, we are not aware of any work doing the same for aesthetics.

Recent works have leveraged the differentiable nature of CNN models to generate ``adversarial examples'', \ie,  examples that have been slightly modified, sometimes in an imperceptible manner, but that lead to completely different, targeted predictions using the original models \cite{szegedy2014intriguing, goodfellow2015explaining}. 
Following those ideas, we assign new distributions to our testing images, which are slightly better or slightly worse than the original predictions of the model. We then modify the images by gradient descent such that the predicted distributions of the modified images better matches those new distributions. The model is kept constant.

Not surprisingly, we found that one could reduce the score of an image by a significantly larger margin and in fewer iterations than one could improve it. However, non-negligible improvements -- up to one point or even more in the mean prediction -- were still possible on most images. However, the resultant changes to the image are visually-imperceptible, both when image scores improve and when they worsen. This suggests that aesthetic models are also somewhat vulnerable to adversarial examples.

\begin{figure}
 \centering
 \includegraphics[width=0.95\linewidth]{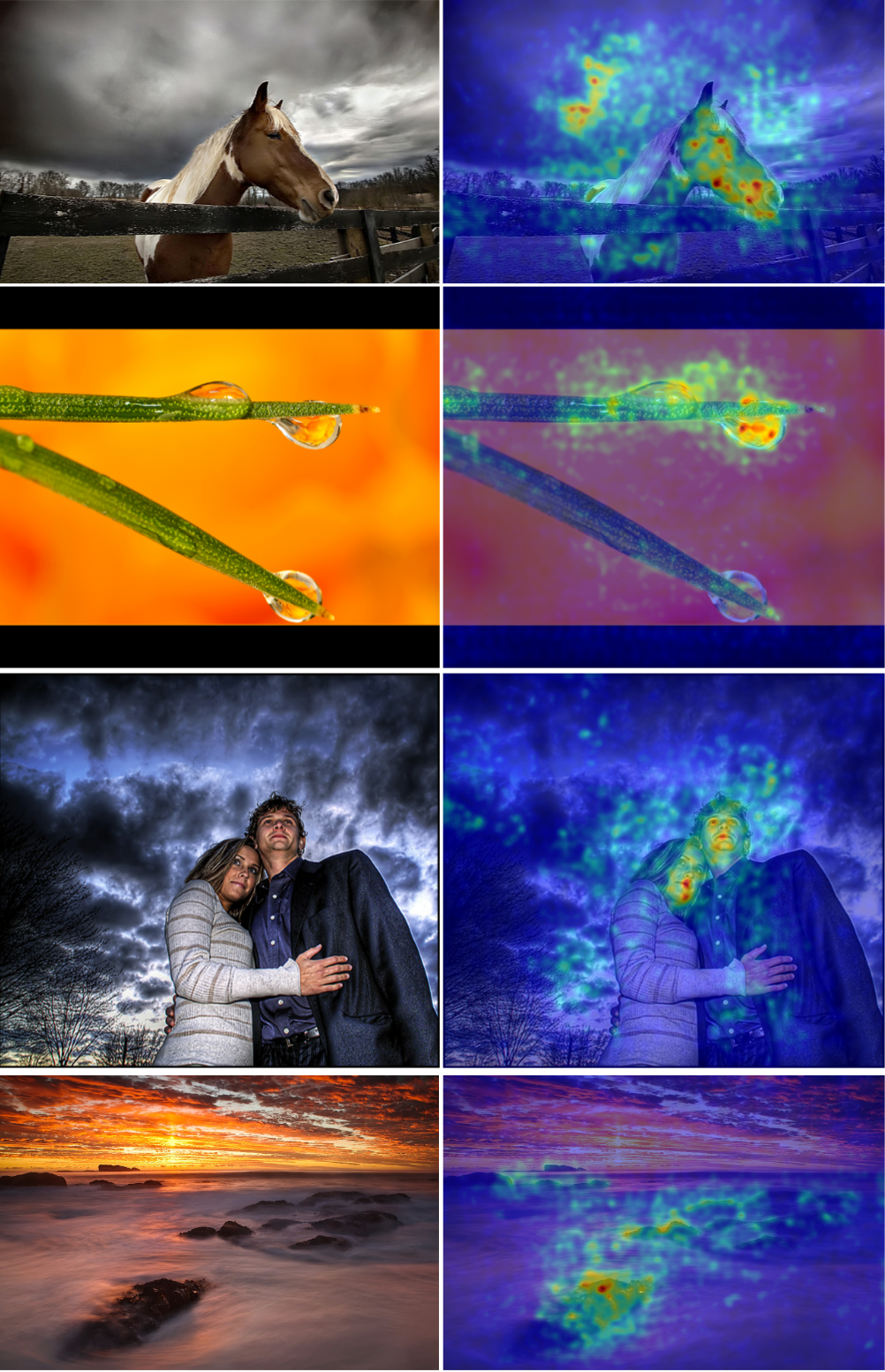}%
 \caption{Adversarial images and their heatmaps, showing which regions of the images were changed to improve or reduce their score. The aesthetic model tends to focus on salient regions.}
 \label{fig:adversarial}  
 \end{figure}

Figure~\ref{fig:adversarial} shows some images from the AVA dataset together with a heatmap of the regions that were modified the most to construct their adversarial examples. We observe changes in background pixels that lead to a loss reduction, but also strong changes in regions that are in focus, contain foreground objects, or contain semantically meaningful content such as faces. This strongly suggests that this type of visual content is used by the model to make predictions.

%% file: conclusion.tex
\section{Conclusions}\label{sec:conclusions}

We proposed an aesthetic prediction model, APM, which gives state-of-the-art performance on three tasks: (i) classification; (ii) score regression; and (iii) score distribution prediction. APM was trained using robust regression techniques using an architecture that can deal with high resolution images without distorting their aspect ratio. Our proposed training procedure leverages semantic information while avoiding the introduction of large auxiliary semantic classification networks.
We also showed how to modify images to improve or worsen their predicted scores. These modified images provided insights into the types of visual content that are often deemed important by the model.
As future work, we would like to leverage knowledge of the most influential image regions for image aesthetics in order to modify images such that their aesthetics are improved.